\documentclass[conference]{IEEEtran}
\usepackage{times}

\usepackage{amsmath}
\usepackage{amsfonts}
\usepackage{amssymb}

\usepackage{graphicx}
\usepackage{multirow}
\usepackage{multicol}

\usepackage{algorithmic}

\usepackage{textcomp}
\usepackage{xcolor}

\usepackage{cite}
\newcommand{\ready}[1]{}
\usepackage[bookmarks=true]{hyperref}

\usepackage[letterpaper]{geometry}
\def\BibTeX{{\rm B\kern-.05em{\sc i\kern-.025em b}\kern-.08em
T\kern-.1667em\lower.7ex\hbox{E}\kern-.125emX}}

\def\ie{{\em {\em i.e.},\ }}

\def\vs{{\em v.s.\/} }

\begin{document}

\title{\vspace{0.25in}Inverse Resistive Force Theory (I-RFT): \\ Learning granular properties through robot-terrain physical interactions}
\author{
  \IEEEauthorblockN{Shipeng Liu$\dagger$,Feng Xue$\dagger$, Yifeng Zhang, Tarunika Ponnusamy, Feifei Qian\textsuperscript{*}}
  \IEEEauthorblockA{University of Southern California, Los Angeles, CA 90089, USA}
  \thanks{\daggerCo-first authors.}
  \thanks{*Corresponding author: feifei@usc.edu}
}
\maketitle
\begin{abstract}
\ready{shipeng +1, FQ+1}
For robots to navigate safely and efficiently on soft, granular terrains, it is crucial to gather information about the terrain’s mechanical properties, which directly affect locomotion performance. Recent research has developed robotic legs that can accurately sense ground reaction forces during locomotion. However, existing tests of granular property estimation often rely on specific foot trajectories, such as vertical penetration or horizontal shear, limiting their applicability of being used during natural locomotion. To address this limitation, we introduce a physics-informed machine learning framework, Inverse Resistive Force Theory (I-RFT), which integrates the Granular Resistive Force Theory model with Gaussian Processes to infer terrain properties from proprioceptively measured contact forces under arbitrary gait trajectories. By embedding the granular force model within the learning process, I-RFT preserves physical consistency while enabling generalization across diverse motion primitives. Experimental results demonstrate that I-RFT accurately estimates terrain properties across multiple gait trajectories and toe shapes. Moreover, we show that the quantified uncertainty over the terrain resistance stress map could enable robots to optimize foot design and gait trajectories for efficient information gathering. This approach establishes a new foundation for data-efficient characterization of complex granular environments and opens new avenues to locomotion strategies that actively adapt gait for autonomous terrain exploration  
\end{abstract}

\IEEEpeerreviewmaketitle

\section{Introduction}\label{sec:intro}\ready{shipeng +1, FQ+1}
\begin{figure}[h]
  \centering
\includegraphics[width=0.44\textwidth]{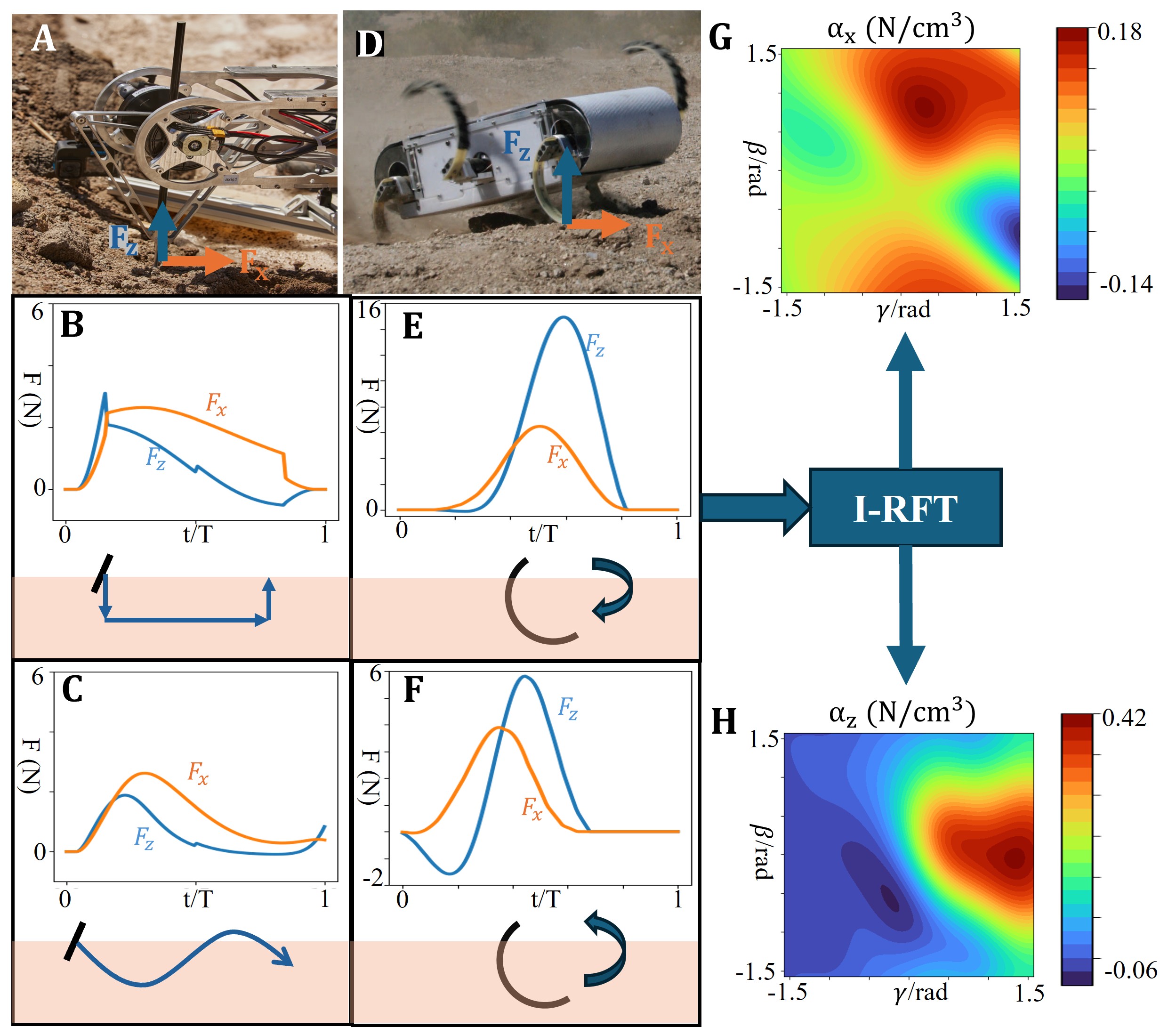}
  \caption{Configuration-dependent robot–terrain interaction forces motivate inverse stress-map inference. (A, D) Two robotic platforms  with variations in leg geometry and trajectory: a direct-drive robotic leg (A), and a C-legged hexapedal robot (D). (B, C) Example interaction forces of the direct-drive robotic leg under rectangular and cubic trajectories, respectively. (E, F) Example forces generated by the C-shaped robot leg during clockwise and counterclockwise rotation. For the same terrain, the measured interaction forces vary significantly with leg morphology and trajectory, demonstrating the configuration-dependent and the difficulty of directly inferring intrinsic terrain properties.
  (G, H) Stress map of the granular terrain in $x$ and $z$ directions, respectively. These intrinsic terrain properties are configuration-independent and are inferred using the proposed I-RFT framework. 
  }
  \label{fig:cover}
\end{figure}

Deformable terrains such as sand, mud, and gravel exist widely in natural environments, including deserts, forests, disaster zones, and planetary surfaces. The capability to robustly traverse these terrains is therefore essential for field robotics and space exploration applications~\cite{doi:10.1126/scirobotics.abc5986, jiang2025safe, Liu2023THRI, Liu2024hri}. The mechanical behavior of these substrates is inherently complex, exhibiting solid-like or fluid-like characteristics that depend on substrate properties such as compaction and cohesion~\cite{shakeel2019rheological,nie2020investigation, liu2025adaptive, liu2023mud}. Such complexity requires robots to reason about substrate properties and mechanical responses in order to adapt locomotion strategies and avoid catastrophic mobility failures \cite{li2009sensitive, mazouchova2013flipper, qian2015principles, qian2013walking, liu2023adaptation, liu2025adaptive, liu2025bio}. 

One key challenge of obtaining accurate substrate property estimation is that, for deformable substrates like sand, mud, and snow,  physical properties such as compaction and cohesion can be difficult to discern using vision-based methods alone~\cite{8823987, ren2024topnavleggednavigationintegrating}. Proprioceptive force measurement, enabled by direct-drive motors with high force transparency~\cite{kenneally2018actuator, Sungbae2016}, offers a promising alternative for terrain property characterization~\cite{liu2026scout, fulcher2025effect}. Recent research has developed direct-drive robots that can use their legs as  rheometers~\cite{ruck2024downslope}, accurately estimating substrate properties, including penetration resistance and shear strength, from a wide range of granular terrain materials~\cite{qian2019rapid,ruck2024downslope}. 
These approaches enable robots to measure deformable substrate properties during leg-ground interaction without the need for external sensors. However, these approaches often rely on specific leg intrusion trajectories, such as vertical penetration and horizontal shear~\cite{qian2019rapid, liu2026scout, liu2025adaptive}, which limit their applicability for incorporation into natural robot locomotion, and constrains the ability to acquire spatially dense terrain estimates during each step. 

To address this challenge, this work proposes Inverse Resistive Force Theory (I-RFT)~\cite{li2013terradynamics,agarwal2023mechanistic}, a physics-informed Gaussian process framework that enables terrain mechanics inference from arbitrary gait trajectories and foot shapes. Rather than learning contact forces directly from data, I-RFT adopts a granular Resistive Force Theory–based formulation in which leg–terrain forces are expressed as the superposition of distributed, orientation-dependent stress contributions along the leg. 
In this representation, the net leg–terrain force arises from integrating local stress contributions over the contact surface, with each contribution determined by the local element orientation and direction of motion relative to the granular medium. These relationships are captured by terrain-specific stress-per-depth maps, which encode how granular resistance varies with intrusion geometry and motion direction~\cite{lichenrft, agarwal2023mechanistic}. By using these stress maps as the latent variables of inference, I-RFT provides a physically meaningful parameterization of terrain mechanics that is compatible with arbitrary gait trajectories and foot geometries.

An additional challenge arises from the indirect and aggregated nature of proprioceptive sensing during robot locomotion. In practice, the available observations are joint-level torque measurements~\cite{liu2025adaptive, fulcher2025effect}, which reflect the combined effects of stress contributions from multiple leg segments simultaneously interacting with the terrain. Because each segment experiences different orientations and motion directions, the underlying terrain stress field is not observed directly but only through aggregated joint-level signals produced by the robot’s kinematics. This indirect observation process violates the pointwise observation assumptions underlying standard Gaussian Process regression and instead requires inference over a latent function from linear functionals of the field, motivating a linear inverse problem formulation.

To address this challenge, I-RFT explicitly couples the RFT forward model with the robot’s proprioceptive sensing pipeline to formulate a structured linear inverse problem~\cite{randrianarisoa2023variational}. Each proprioceptive observation is modeled as a linear functional of the terrain-specific stress-per-depth map, with RFT providing the mapping from local stresses to net contact forces and the robot’s kinematics and Jacobians projecting these forces into joint-level measurements. I-RFT captures this end-to-end sensing process using composite Gaussian Process kernels, enabling the continuous stress map to be inferred directly from aggregated proprioceptive data collected under arbitrary gait trajectories and toe geometries.

Our proposed I-RFT yields stress-map reconstructions with calibrated uncertainty across both C-shape and I-shape legs under diverse trajectories, enabling robots to actively choose gaits and toe geometries for more informative terrain probing. This paper is organized as follows: Section~\ref{sec:prelim} reviews RFT/GP preliminaries that ground our model assumptions. Section~\ref{sec-irft} formulates the inverse RFT framework and its sensing pipeline. Section~\ref{sec:exp} details hardware, trajectories, and the experiments we conducted both numerical and experimental. Section~\ref{sec:results} reports reconstruction accuracy and uncertainty-guided insights. Section~\ref{sec:conclusion} summarizes contributions and future directions.


\section{Preliminaries}
\label{sec:prelim}
\ready{shipeng +1, FQ+1}
\subsection{Resistive Force Theory (RFT)}

\begin{figure}[h]
  \centering
\includegraphics[width=0.45\textwidth]{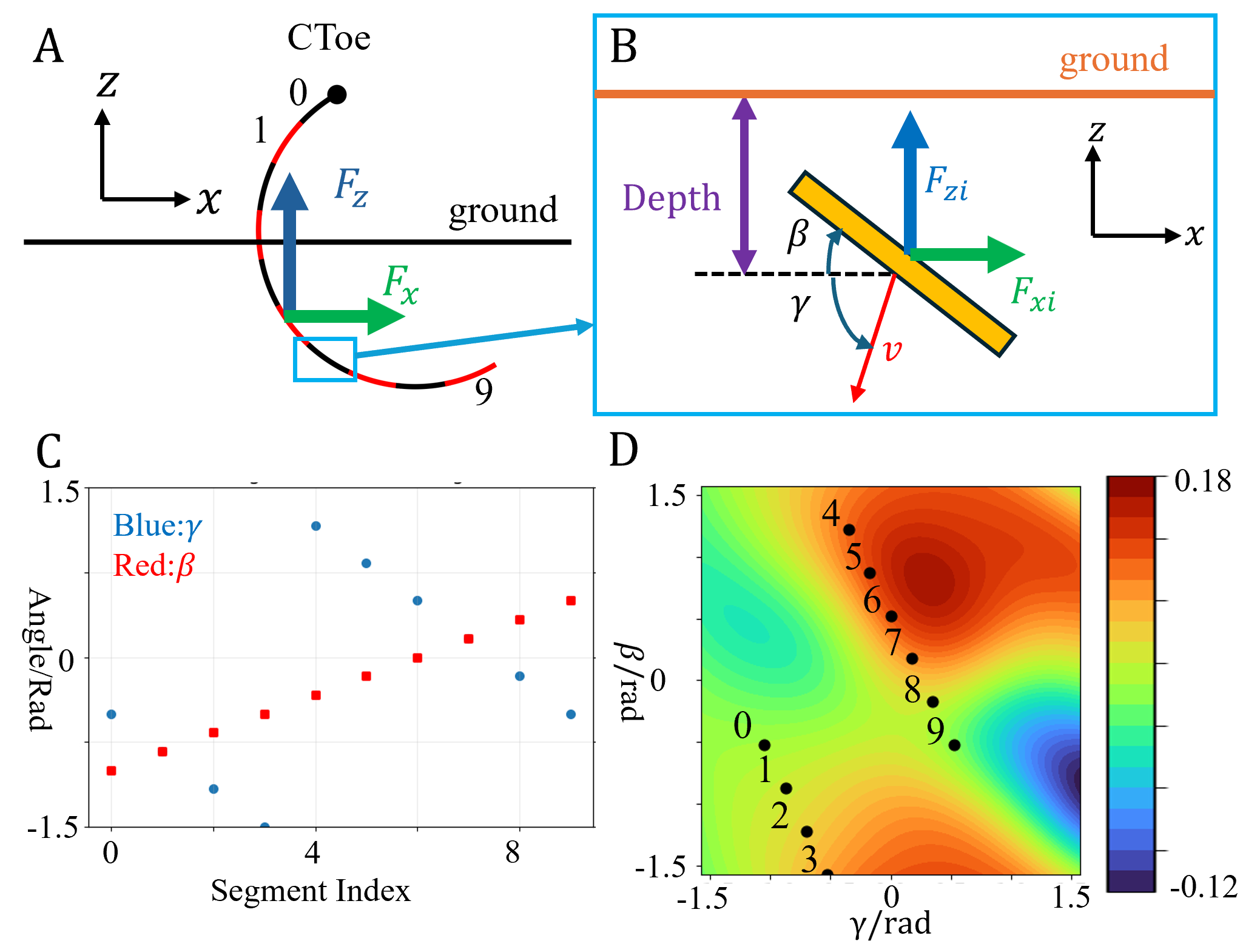}
  \caption{Forward Resistive Force Theory (RFT) formulation. (A) C-Toe discretized into 10 surface segments (labeled with index 0 to 9) for RFT evaluation. (B) Zoomed-in diagram illustrating local geometric parameters defining segment orientation, $\beta$, and motion direction, $\gamma$.  (C) Instantaneous segment angles across the contact surface.. (D) Forward mapping: each segment samples a location in the terrain stress map and contributes to the net force through weighted superposition.}
  \label{fig:rft}
\end{figure}

Resistive Force Theory (RFT) is a reduced-order terradynamics model that approximates the interaction between a locomotor and granular media by integrating local resistive forces acting on differential surface elements~\cite{lichenrft}. According to RFT, in the vertical plane, the force on a surface element depends only on its depth below the surface, its orientation, and its direction of motion. Using this formulation, the granular resistive forces exerted on a limb or footpad can represented as a superposition of forces action on a collection of discrete surface segments with varying depth, orientation, and motion direction. For a discretized limb composed of \( M \) surface segments (Fig. \ref{fig:rft}A), the total lift and drag forces in the vertical plane are given by
\begin{equation}
F_{z,x} = \sum_{m=1}^{M} |z_m|\, A_m \, \alpha_{z,x}(\beta_m, \gamma_m),
\label{eq:rft_discrete}
\end{equation}
where \( |z_m| \) denotes the depth of the \( m \)-th surface segment below the free surface, \( A_m \) is its surface area, \( \beta_m \) is the angle between the surface tangent and the horizontal, and \( \gamma_m \) is the angle between the segment velocity and its surface normal.(Fig.\ref{fig:rft}B) The function \( \alpha_{z,x}(\beta_m, \gamma_m) \) represents the stress per unit depth associated with the segment’s orientation and direction of motion, refer to Fig.\ref{fig:rft}D.



\subsection{Gaussian Process Regression}

Gaussian Processes (GPs) provide a nonparametric Bayesian framework for learning unknown continuous functions. Let \( f : \Theta \subset \mathbb{R}^d \rightarrow \mathbb{R} \) denote a latent function defined over an input domain \( \Theta \). A GP prior over \( f \) is specified as
\[
f(\cdot) \sim \mathcal{GP}\big(m(\cdot), k(\cdot, \cdot)\big),
\]
where \( m(\cdot) \) is the mean function and \( k(\cdot, \cdot) \) is the covariance (kernel) function, encoding prior assumptions on smoothness, periodicity, and correlations. Given a dataset of noisy direct observations $
\mathcal{D} = \{(\theta_i, y_i)\}_{i=1}^t$, $\theta_i \in \Theta$, each observation is assumed to follow $y_i = f(\theta_i) + \varepsilon_i$, $\varepsilon_i \sim \mathcal{N}(0, \sigma_n^2)$, with i.i.d.\ Gaussian noise of variance \( \sigma_n^2 \). Under the GP prior, the training output \( \mathbf{y} = [y_1, \dots, y_t]^\top \) follows a multivariate Gaussian distribution $\mathbf{y} \sim \mathcal{N}\big(\mathbf{m},\, K + \sigma_n^2 I\big)$
where \( \mathbf{m} = [m(\theta_1), \dots, m(\theta_t)]^\top \) and \( K \in \mathbb{R}^{t \times t} \) is the kernel matrix with entries \( K_{ij} = k(\theta_i, \theta_j) \).

For a new test input \( \theta_* \), the posterior predictive distribution of \( f(\theta_*) \) is Gaussian with mean and variance given by
\begin{align}
\mu(\theta_*) &= k(\theta_*, \Theta)\,[K + \sigma_n^2 I]^{-1} \mathbf{y}, \\
\sigma^2(\theta_*) &= k(\theta_*, \theta_*) 
- k(\theta_*, \Theta)\,[K + \sigma_n^2 I]^{-1} k(\Theta, \theta_*).
\end{align}
The predictive mean \( \mu(\theta_*) \) represents the estimated function value, while the predictive variance \( \sigma^2(\theta_*) \) quantifies uncertainty. The kernel hyperparameters, including signal variance, lengthscales, and noise variance, are commonly learned by maximizing the marginal log-likelihood of the observations~\cite{seeger2004gaussian}.

\section{Inverse-RFT}
\label{sec-irft}
\ready{shipeng +1, FQ+1}

\subsection{Problem formulation and Inverse RFT overview}
During robot locomotion, terrain stress maps over local interaction angles $(\beta,\gamma)$ are not directly observable. Instead, we observe only force measurements that aggregate stress contributions from multiple contact segments. To infer terrain mechanical properties from these indirect observations, we formulate the \emph{Inverse RFT} problem: reconstructing terrain stress maps from force observations and robot state.

To address the challenges of (i) modeling forward forces under arbitrary geometries and (ii) handling composite observations, we model latent stress maps as Gaussian processes and leverage RFT to decompose forces into orientation-dependent stress contributions (Section~\ref{sec-probabilistic-modeling}), then derive a posterior predictive distribution for composite observations (Section~\ref{sec-composite}). 

\subsection{Probabilistic Modeling}
\label{sec-probabilistic-modeling}

During locomotion, we observe composite external contact forces in the robot frame, denoted as \(\mathbf{F} = [F_z, F_x]^\top\) (vertical then horizontal). These forces arise from the integral of local stress contributions across all contact segments. Specifically, each force vector \(\mathbf{F}_i = [F_{i,z}, F_{i,x}]^\top\) at time step \(i\) follows:
\begin{align}
\mathbf{F}_i &= \sum_{m=1}^{M} |z_{i,m}| A_{i,m} \begin{bmatrix} \alpha_z(\theta_{i,m}) \\ \alpha_x(\theta_{i,m}) \end{bmatrix} + \boldsymbol{\varepsilon}_i
\label{eq:rft_force_integration}
\end{align}
where $\boldsymbol{\varepsilon}_i \sim \mathcal{N}(\mathbf{0}, \sigma_n^2 I_2)$, \(|z_{i,m}|\) and \(A_{i,m}\) are the known depth and area of the \(m\)-th segment at time step \(i\), \(\theta_{i,m} = (\beta_{i,m}, \gamma_{i,m})\) are the local interaction angles computed from the robot state and contact geometry, and \(\boldsymbol{\varepsilon}_i\) is i.i.d.\ Gaussian noise with variance \(\sigma_n^2\). Each observation \(\mathbf{F}_i\) thus combines evaluations of both \(\alpha_z(\theta)\) and \(\alpha_x(\theta)\) at multiple known angle pairs \(\{\theta_{i,m}\}_{m=1}^{M}\).

Given a dataset comprising multiple composite observations, $
\mathcal{D}_t = \{(\{\theta_{i,m}\}_{m=1}^{M}, \{w_{i,m}\}_{m=1}^{M}, \mathbf{F}_i)\}_{i=1}^{t},
\label{eq:composite_dataset}$, where \(\mathbf{F}_i \in \mathbb{R}^{2}\) is the composite force vector at time step \(i\). The scalar weights \(w_{i,m} = |z_{i,m}| A_{i,m}\) encode the geometric scaling factors that convert stress per unit depth to force contributions: each weight multiplies the stress map evaluation \(\alpha_z(\theta_{i,m})\) or \(\alpha_x(\theta_{i,m})\) to yield the force contribution from segment \(m\) at time step \(i\), as shown in Eq.~\eqref{eq:rft_force_integration}.

To model the latent stress maps \(\alpha_z(\beta, \gamma)\) and \(\alpha_x(\beta, \gamma)\), we place independent Gaussian process priors over them. To model smooth spatial variation in the stress maps, we use a Radial Basis Function (RBF) kernel. Following the original RFT formulation~\cite{lichenrft}, the stress maps exhibit periodic boundary conditions: $\beta$ has periodicity $\pi$ (i.e., $\beta = 0$ and $\beta = \pi$ represent the same physical configuration), while $\gamma = 0$ and $\gamma = \pi$ represent different cases. To enforce the $\pi$-periodicity in $\beta$, we apply a sinusoidal feature embedding, $
\phi(\theta) = (\sin(2\beta), \cos(2\beta), \sin(\gamma), \cos(\gamma))
$, where the factor of 2 in the $\beta$ terms ensures $\pi$-periodicity, and define the kernel as
$
k(\theta,\theta') = \sigma_f^2
\exp\!\left(-\frac{1}{2}\sum_{d=1}^{4}
\frac{(\phi_d(\theta)-\phi_d(\theta'))^2}{\ell_d^2}\right)
$, where $\sigma_f^2$ is the signal variance and $\ell_d$ are dimension-wise lengthscales. 

This formulation represents a \emph{linear inverse problem}~\cite{randrianarisoa2023variational}, where the observations are linear operator evaluations of the latent GP functions. Specifically, each force observation is a linear combination of function evaluations: \(\mathbf{F}_i = \mathcal{L}_i[\alpha_z, \alpha_x] + \boldsymbol{\varepsilon}_i\), where \(\mathcal{L}_i\) is a linear operator that applies the weights \(w_{i,m}\) to evaluate both stress components at multiple angle pairs. The joint probability distribution over these composite observations follows a Gaussian distribution:
\begin{equation}
\mathbf{F}_t \sim \mathcal{N}(\mathbf{0}, W_z K_z W_z^\top + W_x K_x W_x^\top + \sigma_n^2 I),
\label{eq:joint_distribution_composite}
\end{equation}
where the weight matrices $W_i \in \mathbb{R}^{2t \times tM}$ 
and kernel matrices $K_i \in \mathbb{R}^{tM \times tM}$ 
(for $i \in \{x,z\}$) are defined as
\begin{align}
(K_i)_{(p,m),(q,m')} &= k_i(\theta_{p,m}, \theta_{q,m'}), \\
(W_i)_{(p,c),(q,m)} &= w_{p,m}\,\delta_{pq}\,\mathbb{I}[c=i].
\end{align}
Here $p,q \in \{1,\dots,t\}$ index time steps, 
$m,m' \in \{1,\dots,M\}$ index segments, 
and $c \in \{x,z\}$ denotes force components.


\subsection{Posterior Prediction with Composite Observation}
\label{sec-composite}
Given the probabilistic model defined in Section~\ref{sec-probabilistic-modeling}, we now derive the posterior predictive distribution of granular terrain properties for making predictions from composite observations. Given a new test input describing a toe-terrain interaction configuration, characterized by a set of angle pairs $\{\theta_*^{(m)} = (\beta_*^{(m)}, \gamma_*^{(m)})\}_{m=1}^{M}$ describing the orientation and movement direction of each contact segment, along with corresponding known geometric weights $\{w_*^{(m)}\}_{m=1}^{M}$ (computed from segment depths $|z_*^{(m)}|$ and areas $A_*^{(m)}$ as in Eq.~\eqref{eq:rft_force_integration}), we can make predictions for the stress map and external forces:


\textbf{Stress map prediction:} The learned stress map $\alpha_z(\theta)$ (and similarly $\alpha_x(\theta)$) is a continuous function over the angle space $\theta = (\beta, \gamma)$. For any test angle pair $\theta_* = (\beta_*, \gamma_*)$, we can predict the stress map value with uncertainty. Since the composite observations aggregate multiple segments, we must account for the full covariance induced by the observation operator when predicting each component. For the vertical component $\alpha_z$, the posterior mean and variance are:
\begin{align}
\mu_{\alpha_z}(\theta_*) &= \mathbf{k}_z(\theta_*, \Theta)^\top W_z^\top \nonumber \\
&\quad \times (W_z K_z W_z^\top + W_x K_x W_x^\top + \sigma_n^2 I)^{-1} \mathbf{F}_t, \label{eq:stress_z_mean} \\
\sigma_{\alpha_z}^2(\theta_*) &= k_z(\theta_*, \theta_*) - \mathbf{k}_z(\theta_*, \Theta)^\top W_z^\top \nonumber \\
&\quad \times (W_z K_z W_z^\top + W_x K_x W_x^\top + \sigma_n^2 I)^{-1} \nonumber \\
&\quad \times W_z \mathbf{k}_z(\theta_*, \Theta), \label{eq:stress_z_var}
\end{align}here $\mathbf{k}_z(\theta_*, \Theta)$ is a column vector of kernel evaluations $[k_z(\theta_*, \theta_{1,1}), \dots, k_z(\theta_*, \theta_{t,M})]^\top$ between the test angle $\theta_*$ and all training angle pairs, $W_z$ and $K_z$ are the weight and kernel matrices for the vertical component, and $\mathbf{F}_t$ contains all composite force observations. The horizontal component $\alpha_x$ follows the same procedure.

\textbf{Force prediction:} 
The learned stress map enables prediction of future leg–terrain contact forces 
$\mathbf{F}_* = [F_{x,*}, F_{z,*}]^\top$ for arbitrary toe–terrain interactions, 
which is essential for locomotion adaptation under varying terrain properties. 

For each contact configuration, we evaluate the stress map at the segment angle 
$\theta_*^{(m)} = (\beta_*^{(m)}, \gamma_*^{(m)})$, where $m = 1,\dots,M$ indexes 
surface segments with different orientations. Using the RFT forward model 
(Eq.~\eqref{eq:rft_discrete}), the predicted mean horizontal force is
\vspace{-0.1cm}
\begin{align}
F_{x,*} 
= \sum_{m=1}^{M} |z_*^{(m)}| A_*^{(m)} \mu_{\alpha_x}(\theta_*^{(m)}),
\end{align}
\vspace{-0.1cm}
where $\mu_{\alpha_x}(\theta_*^{(m)})$ denotes the GP posterior mean of the 
horizontal stress component. The vertical component $F_{z,*}$ is computed 
analogously using $\mu_{\alpha_z}(\theta_*^{(m)})$. Uncertainty in force prediction propagates from the stress map uncertainty. 
Assuming independence between segments, the predictive variance of the 
horizontal force is $
\sigma_{F_x}^2 
= \sum_{m=1}^{M} \left(|z_*^{(m)}| A_*^{(m)}\right)^2 
\sigma_{\alpha_x}^2(\theta_*^{(m)})
$ and $\sigma_{F_z}^2$ is obtained analogously.

With composite observations, the marginal log-likelihood of the force 
observations $\mathbf{F}_t \in \mathbb{R}^{2t}$ becomes
\vspace{-0.1cm}
\begin{align}
\log p(\mathbf{F}_t)
&= -\frac{1}{2}\mathbf{F}_t^\top 
\left( W_z K_z W_z^\top + W_x K_x W_x^\top + \sigma_n^2 I \right)^{-1} 
\mathbf{F}_t \nonumber \\
&\quad - \frac{1}{2} \log 
\left| W_z K_z W_z^\top + W_x K_x W_x^\top + \sigma_n^2 I \right|  \nonumber \\
&\quad - \frac{tN}{2} \log(2\pi).
\label{eq:marginal_log_likelihood}
\end{align}

\section{Results} 
\label{sec:results}
We conduct both numerical simulations and real-world experiments to validate the effectiveness of I-RFT. Sec.~\ref{sec:numerical} evaluates the accuracy of reconstructed stress maps across two toe geometries and multiple gait trajectories, and examines the effects of toe geometry, trajectory design, and measurement noise on terrain property inference performance. Sec.~\ref{sec:exp} further demonstrates the applicability of the method using a physical robotic leg platform.

\subsection{Numerical Validation}\label{sec:numerical}
\ready{ shipeng +1, FQ+1}
For numerical validation, we use forward RFT (Eq.~\ref{eq:rft_discrete}) together with the reported stress maps from~\cite{lichenrft} to generate synthetic force measurements for predefined toe geometries and trajectories. Given a toe geometry and trajectory, we compute the contact angles and penetration depths of each discretized segment, then apply Eq.~\ref{eq:rft_discrete} to obtain the resulting net force. By using Eq. \ref{eq:rft_discrete}, we can obtain the corresponding force that acts on the entire robot toe. In simulation, we assume perfect knowledge of robot state (segment angles and depths) and inject Gaussian noise into force observations to emulate measurement uncertainty. The reconstructed stress maps from I-RFT are then compared against the ground-truth maps used in the forward model.

\begin{figure*}[t]
    \centering
    \includegraphics[width=0.9\linewidth]{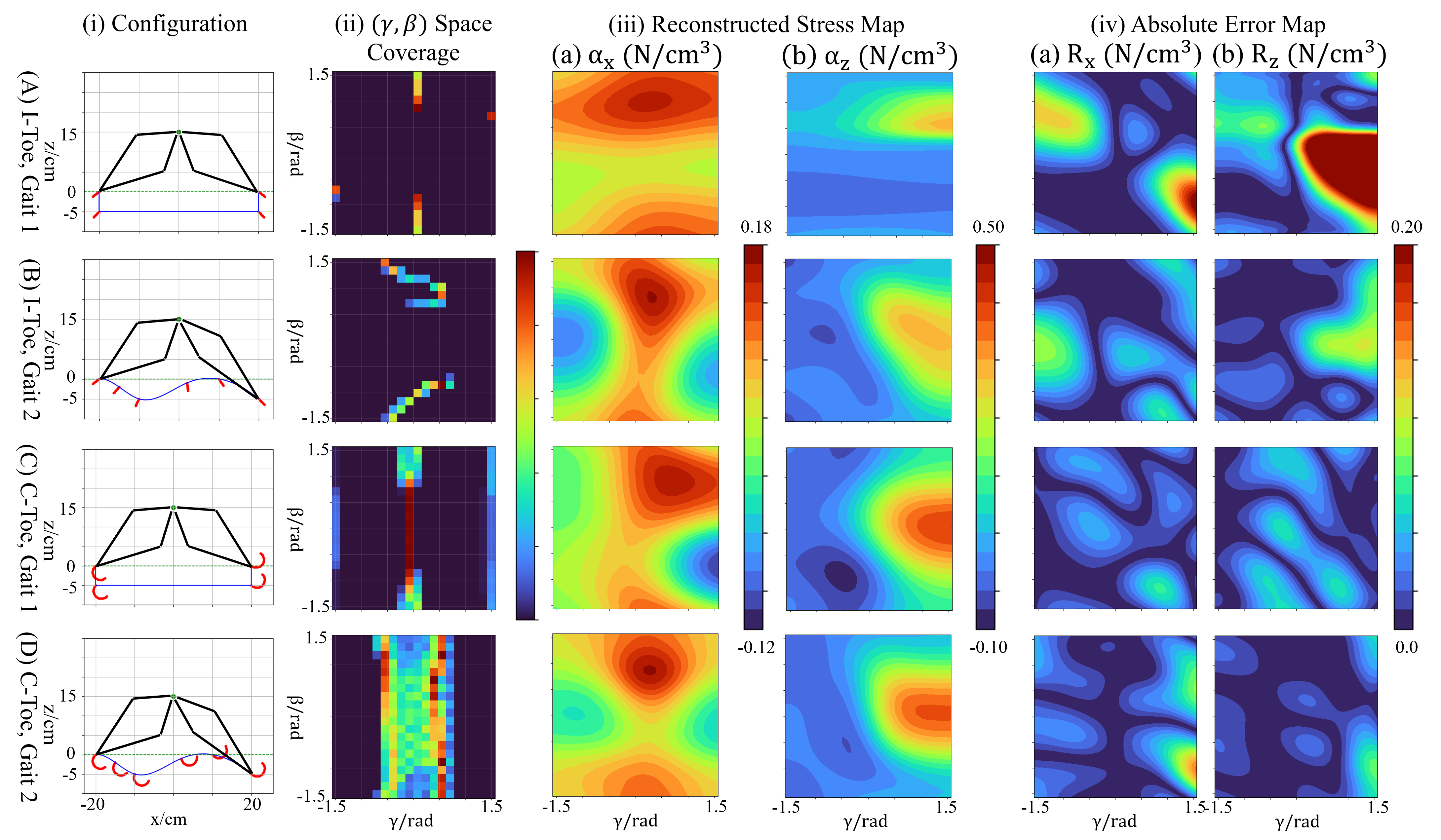}
    \caption{Numerical validation of I-RFT across toe geometries and trajectories for the four configurations: (A) I-Toe, Gait 1; (B) I-Toe, Gait 2; (C) C-Toe, Gait 3; (D) C-Toe, Gait 2. (i) Experimental configurations: blue curve is toe trajectory; red curve indicates toe geometry; dashed line denotes sand surface. (ii) Sampling coverage in the ($\beta$, $\gamma$) domain induced by each configuration. (iii) Reconstructed stress maps, ($\alpha_x$, $\alpha_z$). (iv) Reconstruction error relative to ground truth. }
    \label{fig:simu_res}
\end{figure*}

To evaluate the performance and generalization capability of the proposed I-RFT method across different contact geometries, we test two toe geometry primitives: a I-Toe (flat plate, 20 mm length, 8 mm width, Fig.~\ref{fig:simu_res}i-A, B), and a C-Toe (semicircular, 20 mm radius, 8 mm width, Fig.~\ref{fig:simu_res}i-C, D). The I-Toe segments share identical orientation ($\beta$), yielding limited angular diversity. In contrast, the C-Toe spans the full $\beta$ range along its curved profile, and segment velocities vary across the contact surface, producing a richer sampling of the ($\beta$, $\gamma$) domain. 

For each toe configuration, we test two trajectories: a rectangle trajectory (Fig.~\ref{fig:simu_res}i-A, C), with 50 mm penetration, 400 mm shear, and 50 mm extraction; and a cubic spline trajectory (Fig.~\ref{fig:simu_res}i-B, D),  defined using four control points: (-20, 0), (-10, -5), (10, 0), and (20, -5) (cm). 
The cubic trajectory produces smoother and more continuous variation in $\gamma$, including both positive and negative motion directions. Together, these configurations allow us to isolate the effects of contact geometry and motion design on stress-map reconstruction.

To quantitatively evaluate stress-map reconstruction, we use error-based metrics (RMSE, MAE and ACR$\%$) and correlation/fit metrics ($R^2$ and Pearson correlation). RMSE emphasizes large local discrepancies and is therefore sensitive to outliers, while MAE provides a measure of the absolute deviation. ACR$\%$ indicates area fraction of reconstructed map where error falls within 5\% of the ground truth range. In addition, $R^2$ summarizes how well the estimate explains the variance of the ground truth, and Pearson correlation assesses linear agreement in spatial trends. Together, these metrics capture both absolute accuracy and pattern consistency between the estimated and ground-truth stress maps.

Across all configurations, I-RFT successfully recovers the dominant low-frequency spatial structure of the terrain stress maps (Fig.~\ref{fig:simu_res} iii). RMSE remains below 10\% for most configurations, and correlation metrics ($R^2$, Pearson) indicate strong agreement with ground truth.  However, reconstruction quality varies depending on geometric and trajectory-induced sampling coverage. We analyze these effects below.

\subsubsection{Influence of Toe Geometry}

Toe geometry significantly affects reconstruction accuracy (Tab. \ref{tab:metrics_by_config_axis}). Across both trajectories, the C-Toe consistently outperforms the I-Toe. For example, under Gait 1 ($z$-axis), the C-Toe RMSE is 0.0305, compared to 0.1301 for the I-Toe, and the ACR\% increases from 35.28\% (I-Toe) to 54.86\% (C-Toe). Similarly, correlation metrics ($R^2$, Pearson) are substantially higher for the C-Toe across both axes. These differences are also visible in Fig.~\ref{fig:simu_res}iii, where the C-Toe reconstruction more faithfully recovers both stress magnitude and spatial structure.

This performance gap arises from differences in sampling coverage within the $(\beta,\gamma)$ domain. After discretization, all segments of the I-Toe share the same orientation and velocity direction, so they yield the same $\beta$ and $\gamma$ across samples, leading to highly redundant and spatially concentrated samples (Fig.~\ref{fig:simu_res}ii-A,B). As a result, large regions of the stress map remain weakly constrained, which explains the higher reconstruction error observed in the unsampled areas.
In contrast, the C-Toe consists of segments with varying orientations that span the full range along the curved profile, and each segment’s velocity direction also differs slightly. 
As a result, under Gait 1, the C-Toe covers approximately 10× the $(\beta, \gamma)$ area of the I-Toe, and around 18.1× under Gait 2 than I-Toe (Fig. \ref{fig:simu_res}ii, C, D) and thus provides richer measurements for stress-map reconstruction. 

\subsubsection{Influence of Trajectory}

The trajectory also influence the reconstruction quality, as it directly determines the velocity direction of each segment and thus the resulting sampling coverage in the $(\beta, \gamma)$ domain. Across different trajectories, those whose directions span a wider range of angles can cover a broader region of the parameter space and thus achieve better performance. For example, for the I-Toe, the cubic trajectory allows sample a broader range of the ($\beta$, $\gamma$) space (Fig. \ref{fig:simu_res}ii-A, B). As a result, the cubic trajectory exhibited a substantially improved performance relative to the rectangle trajectory, with a lower RMSE (0.0385 \vs 0.1301) and higher ACR (66.20\% \vs 35.28\%). 

In addition, the sampling region plays a crucial role in reconstruction quality. In particular, the stress map in the X direction exhibits a clear sign transition when $\gamma > \pi/3$, where the resistive force reverses direction as $\beta$ varies. Accurately capturing this transition is essential for reconstructing the shear stress field.
As an example, for the C-Toe under Gait 1, the sampling trajectory passes through this critical region (Fig. \ref{fig:simu_res}ii-C), enabling the model to observe and constrain the sign change in the stress distribution. In contrast, although Gait 2 spans around 5.1× larger total $(\beta, \gamma)$ area overall, it provides less effective coverage within this sign-changing region  (Fig. \ref{fig:simu_res}ii-D), resulting in worse reconstruction performance compared to Gait 1 ($x$-axis RMSE: 0.0332 \vs 0.0236; ACR: 47.18\% \vs  49.67\%).
These results indicate that coverage of physically informative regions is more important than total sampling area alone.

\subsubsection{Noise Sensitivity}
The results reported above are obtained under the assumption that the force measurements are accurate and noise-free. In practice, force sensor measurements are noisy, and factors such as mechanical vibrations can substantially increase the measurement error. To emulate these effects, we add Gaussian noise of varying magnitudes to the forces computed by RFT in the numerical simulation experiments described above, and examine how this affects reconstruction performance.

Evaluation shows that the estimation error increases monotonically with noise level (Table.~\ref{tab:noise_mae_rmse_by_config_axis}). Over the full domain, $\beta,\gamma \in [-\pi/2,\pi/2]$, normalized RMSE increases by 4.12\% for the I-Toe, and  15.85\% for the C-Toe. 
Thus, although the C-Toe performs better under low-noise conditions, it exhibits greater sensitivity to measurement noise.

We hypothesize that the higher noise sensitivity of the C-Toe arises from stronger coupling in the inverse problem. Each force measurement is a weighted sum of stress contributions from a larger set of ($\beta,\gamma$), leading to a more strongly coupled observation. If part of the stress-map value is perturbed, other region's value must compensate to satisfy the constraint. As a result, the solution may exhibit large positive and negative lobes that “balance out” to match the integral equation. Consequently, noise in a localized region can propagate globally across the stress map. 

To test this hypothesis, we define the redundancy ratio as
\begin{equation}
    Redundancy = 1 - \frac{N_{unique}}{N_{total}}
\end{equation}
where $N_{\mathrm{unique}}$ is the number of distinct samples, and $N_{\mathrm{total}}$ is the total number of samples. A higher redundancy value indicates stronger overlap among samples and reduced excitation diversity in the $(\beta, \gamma)$ domain.

Consistent with our hypothesis, when redundancy is high, (See Fig. \ref{fig:simu_res}ii-A, C, Redundancy = 0.97 and 0.68, respectively), repeated constraints in the same region help attenuate the noise, and yields a smooth estimate within the sampled region. However, regions that are weakly sampled or unobserved are dominated by kernel-based extrapolation and reversion to the prior mean, which can lead to over-smoothing and loss of fine-scale (high-frequency) structures. 
In contrast, when redundancy is low (Fig. \ref{fig:simu_res}ii-D, Redundancy = 0.11), the constraints become less overlapping and the posterior may overfit to noisy observations, resulting in high-frequency fluctuations that are inconsistent with the inherently smooth stress map, thereby increasing reconstruction errors. 

Overall, these results reveal a trade-off between excitation diversity and noise robustness: broader angular coverage improves identifiability under low noise, but increases coupling and sensitivity when measurements are noisy.

\begin{table}[ht]
\centering
\small
\renewcommand{\arraystretch}{1.12}
\resizebox{\columnwidth}{!}{%
\resizebox{\columnwidth}{!}{%
\begin{tabular}{l c|ccccc}
\hline
Configuration & Axis & RMSE$\downarrow$ & MAE$\downarrow$ & $R^2$$\uparrow$ & Pearson$\uparrow$ & ACR\%$\uparrow$ \\
\hline
\multirow{2}{*}{I-Toe gait 1}
& X & 0.0552 & 0.0356 & 0.1652 & 0.6757 & 49.59 \\
& Z & 0.1301 & 0.0797 & 0.0525 & 0.4348 & 35.28 \\
\hline
\multirow{2}{*}{I-Toe gait 2}
& X & 0.0369 & 0.0275 & 0.6271 & 0.8562 & 41.72 \\
& Z & 0.0385 & 0.0254 & 0.9169 & 0.9841 & 66.20 \\
\hline
\multirow{2}{*}{C-Toe gait 1}
& X & 0.0236 & 0.0186 & 0.8477 & 0.9246 & 49.67 \\
& Z & 0.0305 & 0.0242 & 0.9480 & 0.9753 & 54.86 \\
\hline
\multirow{2}{*}{C-Toe gait 2}
& X & 0.0332 & 0.0232 & 0.6983 & 0.8364 & 47.18 \\
& Z & 0.0201 & 0.0133 & 0.9772 & 0.9906 & 87.05 \\
\hline
\end{tabular}%
}%
}
\vspace{0.1cm}
\caption{Summary metrics per configuration and axis. The $\downarrow$ and $\uparrow$ indicate lower and higher value signify better performance.}
\label{tab:metrics_by_config_axis}
\end{table}


\begin{table}[t]
\centering
\small
\renewcommand{\arraystretch}{1.12}
\resizebox{\columnwidth}{!}{%
\begin{tabular}{l c|cc}
\hline
Configuration & Axis & RMSE\%$\downarrow$ (0/ 0.05/ 0.2) & MAE\%$\downarrow$ (0/ 0.05/ 0.2) \\
\hline

\multirow{2}{*}{I-Toe gait 1}
& X & 18.26/ 18.41/ 19.33 & 11.63/ 12.42/ 13.40 \\
& Z & 28.97/ 28.11/ 22.65 & 17.30/ 20.71/ 16.95 \\
\hline

\multirow{2}{*}{I-Toe gait 2}
& X & 11.79/ 14.42/ 15.31 &  8.79/ 10.81/ 11.56 \\
& Z &  8.89/ 10.03/ 16.62 &  5.73/ 6.04/ 10.76 \\
\hline

\multirow{2}{*}{C-Toe gait 1}
& X &  6.95/ 9.36/ 25.42 &  5.49/ 7.61/ 21.21 \\
& Z &  7.53/ 11.95/ 16.97 &  5.79/ 9.75/ 12.44 \\
\hline

\multirow{2}{*}{C-Toe gait 2}
& X & 10.72/ 12.86/ 14.55 &  7.64/ 9.44/ 10.54 \\
& Z &  3.68/ 16.18/ 35.37 &  2.41/ 13.50/ 23.11 \\
\hline

\end{tabular}%
}
\vspace{0.1cm}
\caption{Normalized MAE and RMSE under multiplicative Gaussian noise levels 0, 0.05, and 0.2. Noise level is variance of the scale factor with mean value 1. 
}
\label{tab:noise_mae_rmse_by_config_axis}
\end{table}


\subsection{Experimental Validation}\ready{shipeng +1 }
\label{sec:exp}

To demonstrate the applicability of I-RFT on a physical platform, we performed laboratory experiments using a direct-drive robotic leg interacting with a fluidized granular bed (Fig.~\ref{fig:traveler}A). Similarly to numerical simulation, we evaluate two toe geometries: I-Toe (Fig.~\ref{fig:traveler}B, 80 mm length, 8 mm width), and C-Toe (Fig.~\ref{fig:traveler}C, 40 mm radius, 8 mm width); and two trajectories: a rectangle trajectory (Fig.~\ref{fig:traveler}D, 50 mm penetration, 300 mm shear, and 50 mm extraction), and cubic spline trajectory (Fig.~\ref{fig:traveler}E) defined by four control points: (-15cm, 5cm), (-6cm, 0cm), (6cm, 3cm), (15cm, 1cm).  These configurations allow us to assess the robustness and generalization of I-RFT under varying contact geometries and motion patterns in real-world conditions.

\subsubsection{Experimental setup}
\begin{figure}[h]
  \centering
\includegraphics[width=0.45\textwidth]{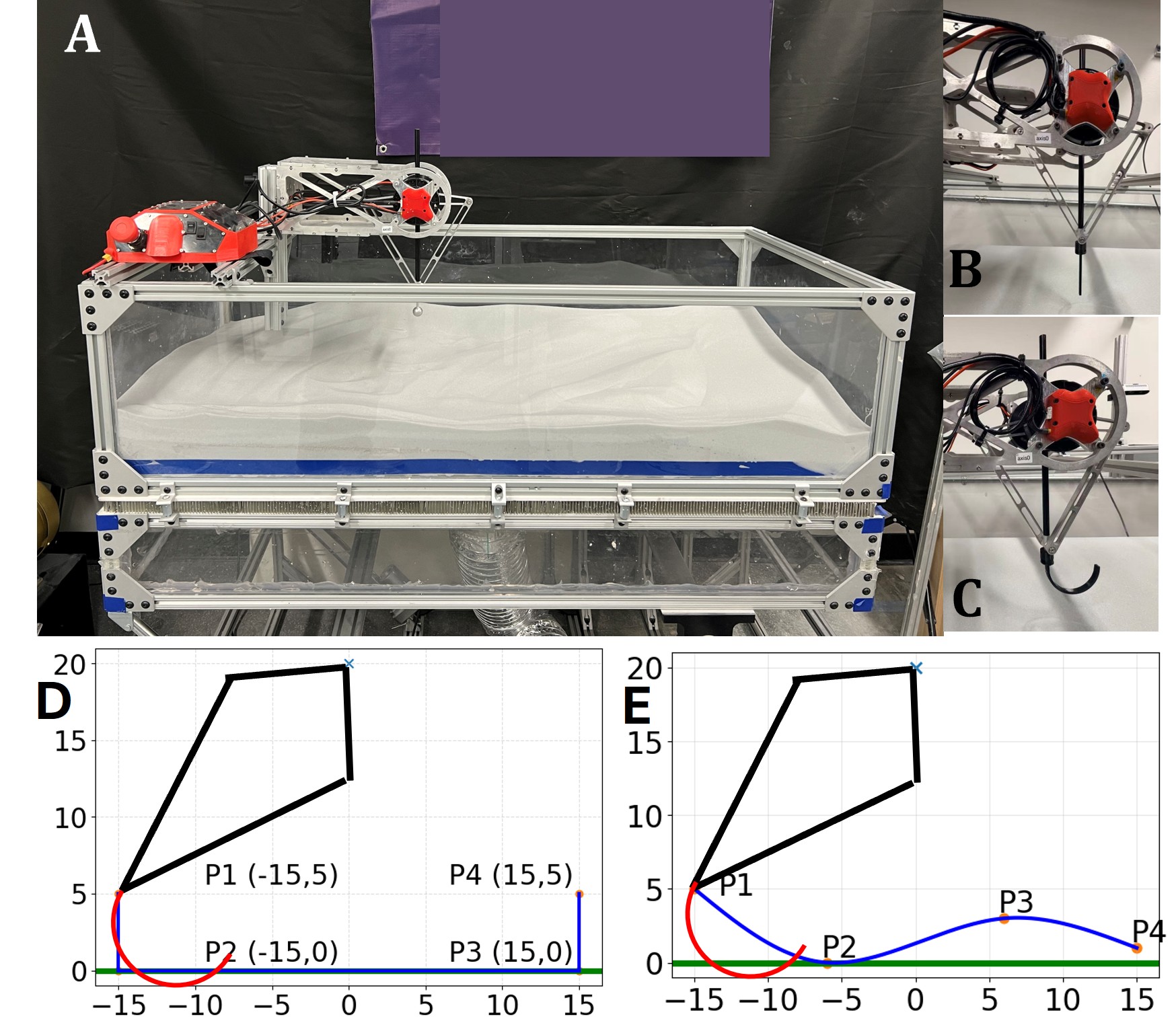}
  \caption{Experimental setup and trajectory configurations. (A) Direct-drive robotic leg mounted on the fluidized granular testbed. (B, C) I-Toe and C-Toe geometries installed on the direct-drive robotic leg. (D) A diagram illustrating gait 1, a rectangular trajectory. (E) A diagram illustrating gait 2, a cubic trajectory. Blue curve represents toe endpoint trajectory; orange dots represents spline control points. Green line represents sand surface level. }
  \label{fig:traveler}
\end{figure}

The robotic leg (Fig.~\ref{fig:traveler}B) consists of a five-bar parallel linkage~\cite{kenneally2016design} actuated by two direct-drive (\ie gearless) motors (U8-II, torque constant $k_t = 0.0973~\mathrm{Nm/A}$). The five-bar architecture enables the toe to execute arbitrary planar (2D) trajectories within its workspace (Fig.~\ref{fig:traveler}D,~\ref{fig:traveler}E).
The direct-drive motors offer high force transparency~\cite{kenneally2018actuator}, allowing estimation of joint torques from measured motor currents:
\begin{equation}
\boldsymbol{\tau} = 
\begin{bmatrix}
\tau_1 \\
\tau_2
\end{bmatrix}
=
k_t
\begin{bmatrix}
I_1 \\
I_2
\end{bmatrix}.
\end{equation}

Under quasi-static operation at a constant linear velocity of $0.02~\mathrm{m/s}$, dynamic effects are negligible. The Cartesian contact force acting on the toe is therefore obtained from the joint torques via the inverse Jacobian transformation:
$
\mathbf{F}_{\text{external}} = \mathbf{J}^{-\top} \boldsymbol{\tau}
\label{eq:jacobian_transform}
$, 
where $\mathbf{F}_{\text{external}} = [f_x, f_z]^\top$ denotes the contact force in the robot frame and $\mathbf{J}$ is the kinematic Jacobian of the leg.

To ensure controlled and repeatable granular terrain conditions, experiments are conducted in a sand fluidized bed (Fig.~\ref{fig:traveler}A, $48 \times 24$~inch), which fluidizes the granular medium by forcing air upward through the particles~\cite{qian2013automated,jin2019preparation}. The medium consists of 300~$\mu$m glass beads filled to a depth of 120~mm. By setting the blower frequency to 60~Hz then slowly ramping down to 0~Hz, the sand can be fully fluidized and then allowed to settle to a flat, consistent compaction state before each trial. 
The robotic leg is rigidly fixed during experiments to isolate leg–terrain interaction dynamics, and the motor-to-ground distance is kept constant throughout.

To eliminate force variations arising from workspace-dependent dynamics, we compute the sand resistive force as $F = F_{\text{raw}} - F_{\text{air}}$ (Fig.~\ref{fig:fieldRES}i, blue curve), where $F_{\text{raw}}$ represents the proprioceptively measured toe force as the leg executes a specific trajectory in the sand, and $F_{\text{air}}$ represents the proprioceptively measured toe force as the toe executes the same trajectory in air. Both $F_{\text{air}}$ and $F_{\text{raw}}$ were passed through a moving-average filter to remove high-frequency oscillations due to joint stiction, and averaged over three trials for repeatability. 

\subsubsection{Experimental results}
\begin{figure}[h]
  \centering
\includegraphics[width=0.48\textwidth]{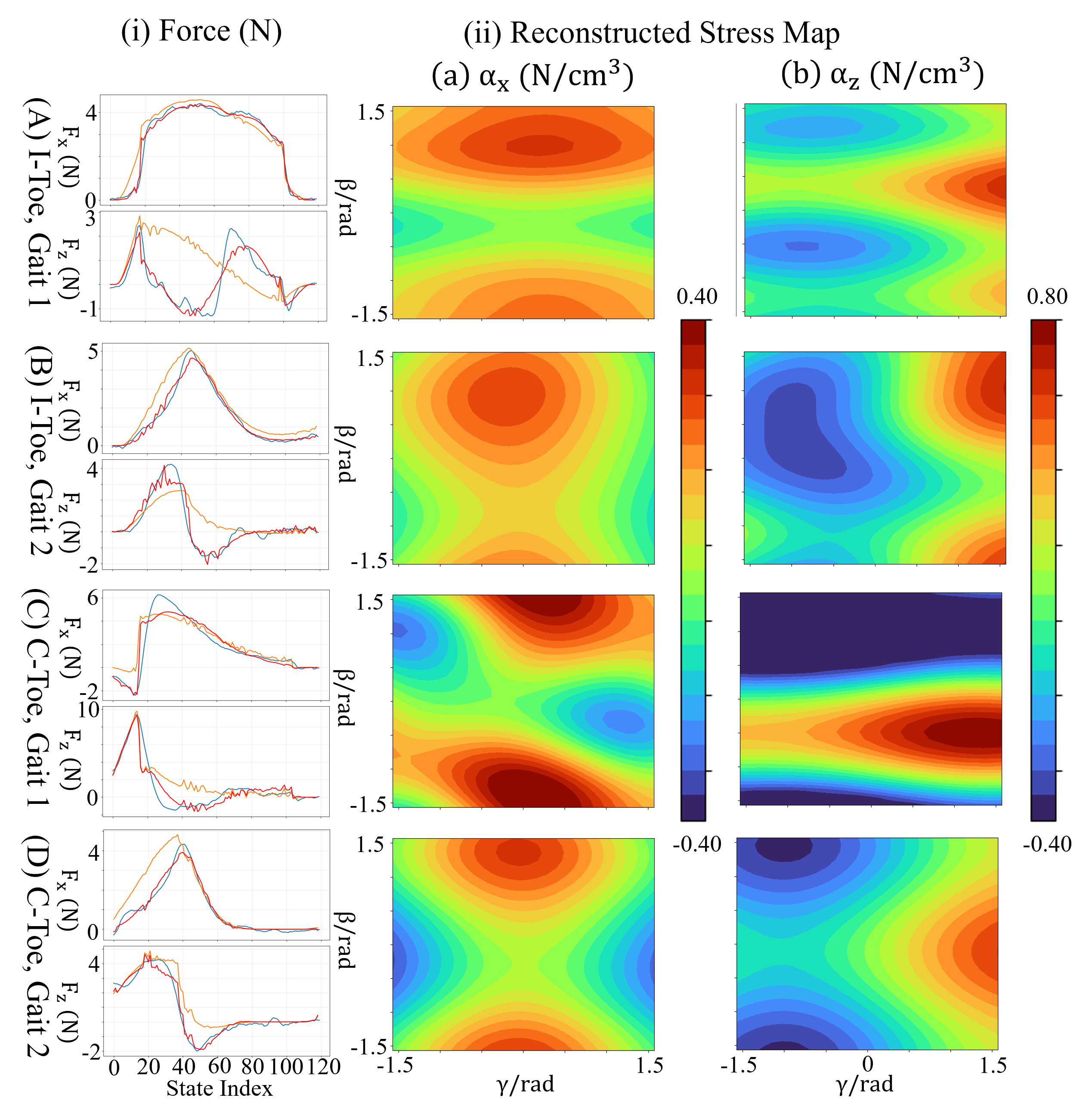}
  \caption{Experimental evaluation of I-RFT. (A–B) I-Toe; (C–D) C-Toe.
(i) Measured sand resistive force (blue), reference RFT prediction using scaled canonical stress maps (orange), and I-RFT reconstruction (red).
(ii) Reconstructed stress maps from experimental data: (a) $\alpha_x$, (b) $\alpha_z$.   }
  \label{fig:fieldRES}
  
\end{figure}

Using the experimentally-sensed sand resistive force (Fig. \ref{fig:fieldRES}i, blue curve), I-RFT robustly extracts the dominant low-frequency characteristics and overall spatial trends of the stress distribution (Fig. \ref{fig:fieldRES}ii). 
To quantitatively evaluate the accuracy of the estimated stress map, we compare the I-RFT–predicted force (Fig.~\ref{fig:fieldRES}i, red curve), obtained using the inferred terrain stress map, with a reference ground-truth force (Fig.~\ref{fig:fieldRES}i, orange curve). Prior granular physics research has shown that angle-dependent granular stress profiles share a generic shape across different granular media and differ primarily by a single medium-dependent scaling factor~\cite{li2013terradynamics}. Following this observation, we construct a reference stress map by scaling the canonical stress profile reported in~\cite{li2013terradynamics}. The scaling factors are determined to be 1.0 in the $x$ direction and 1.5 in the $z$ direction, based on calibration measurements at $(\beta, \gamma) = (pi/3, \pi/2)$. The resulting scaled stress map is then used within the forward RFT model to generate the reference (ground-truth) force profile shown in Fig.~\ref{fig:fieldRES}i, orange curve.

The reconstructed force agreed well for $F_x$, with normalized RMSE values of 21.76\%, 20.02\%, 30.21\%, and 33.77\% across the four configurations. However, the reconstructed $F_z$ exhibited larger errors with normalized RMSE values of 26.98\%, 60.84\%, 114.2\% and 38.68\% across configurations, as compared with the noise-free numerical simulations.  

This reconstruction error mainly arises from two sources. The first is the proprioceptive force sensing accuracy. In practice, the measured toe force is affected by motor friction and cogging effects, as well as by the simplification of modeling the entire toe as a single rigid contact point. As a result, the estimated force does not always follow an ideal Gaussian noise assumption and may exhibit structured or direction-dependent deviations. These non-ideal measurement characteristics introduce bias into the reconstructed stress map. For example, as shown in Fig. \ref{fig:fieldRES}A-i, Fz, when the toe velocity direction becomes nearly horizontal (between state 20 and 100), the measured Z-direction force significantly deviates from the RFT prediction due to hardware sensitivity limitations with RMSE 0.1231.

The second source of error lies in the inherent limitations of the RFT model itself. RFT relies on several simplifying assumptions, including (1) negligible granular flow induced by leg movement~\cite{aguilar2016robophysical}, and (2) a locally flat sand surface. When these assumptions are violated, the predicted forces can deviate systematically from the true interaction forces. For example, as shown Fig. \ref{fig:fieldRES}B-i, D-i, when the toe lift up (between state 40 and 80), the measured Z-direction force significantly deviates from the RFT prediction due to unmodeled sand flow on top of the plate. This leads to substantial reconstruction errors (RMSE 0.2773 and 0.1764 for the $x$ and $z$ directions, respectively). These force constraints directly affect corresponding sampled region ($\gamma$ around and smaller than 0) and propagates inconsistencies to neighboring areas (Fig. \ref{fig:fieldRES} B-ii.b, D-ii.b, large negative values in $\gamma < 0$).

Overall, these experiments validate that I-RFT can recover the dominant structure and spatial trends of granular terrain stress maps from real-world proprioceptive force measurements, despite sensing noise and modeling limitations. The remaining errors primarily stem from non-ideal force estimation and simplifying assumptions in the forward RFT model. Improving force accuracy through better motor calibration, friction compensation, and refined contact modeling should further reduce bias and enhance reconstruction performance.

\section{Conclusion} 
\label{sec:conclusion}

This work introduces Inverse Resistive Force Theory (I-RFT), a physics-informed probabilistic framework that enables robots to infer granular terrain mechanics directly from proprioceptive interaction forces during locomotion. By embedding the forward Resistive Force Theory model within a structured Gaussian Process formulation, I-RFT transforms joint-level torque measurements into spatial terrain stress maps with calibrated uncertainty. Unlike prior methods that rely on prescribed probing motions, our approach generalizes across arbitrary toe geometries and gait trajectories, enabling terrain characterization during natural locomotion. We validate I-RFT through simulations and physical experiments, evaluating reconstruction quality across leg geometries, gait trajectories, and noisy force measurements. These results establish a foundation for interaction-driven perception, where robots infer environmental mechanics through physical contact. Future work will integrate active trajectory optimization and improved force estimation to further close the loop between sensing, inference, and adaptive locomotion.





\bibliographystyle{IEEEtran}
\bibliography{references}

\end{document}